\documentclass[runningheads]{llncs}
\usepackage[T1]{fontenc}
\usepackage{graphicx,verbatim}
\usepackage{amsmath,amssymb}
\usepackage{hyperref}
\usepackage{cleveref}
\newcommand{\myparagraph}[1]{\smallskip\noindent\textbf{#1}}
\usepackage{xcolor}

\crefname{figure}{Fig.}{Figs.}
\Crefname{figure}{Fig.}{Figs.}
\begin{document}
\title{Structure-Guided Histopathology Synthesis via Dual-LoRA Diffusion}
%\titlerunning{Abbreviated paper title}
% If the paper title is too long for the running head, you can set
% an abbreviated paper title here
%
\begin{comment}  %% Removed for anonymized MICCAI submission
\author{First Author\inst{1}\orcidID{0000-1111-2222-3333} \and
Second Author\inst{2,3}\orcidID{1111-2222-3333-4444} \and
Third Author\inst{3}\orcidID{2222--3333-4444-5555}}
%
\authorrunning{F. Author et al.}
% First names are abbreviated in the running head.
% If there are more than two authors, 'et al.' is used.
%
\institute{Princeton University, Princeton NJ 08544, USA \and
Springer Heidelberg, Tiergartenstr. 17, 69121 Heidelberg, Germany
\email{lncs@springer.com}\\
\url{http://www.springer.com/gp/computer-science/lncs} \and
ABC Institute, Rupert-Karls-University Heidelberg, Heidelberg, Germany\\
\email{\{abc,lncs\}@uni-heidelberg.de}}

\end{comment}

\author{Xuan Xu \and Prateek Prasanna}

\authorrunning{Xu and Prasanna}

\institute{Stony Brook University \\
\email{xuanxu@cs.stonybrook.edu, prateek.prasanna@stonybrook.edu}}
  
\maketitle              % typeset the header of the contribution
\begin{abstract}

Histopathology image synthesis plays an important role in tissue restoration, data augmentation, and modeling of tumor microenvironments. 
However, existing generative methods typically address restoration and generation as separate tasks, although both share the same objective of structure-consistent tissue synthesis under varying degrees of missingness, and often rely on weak or inconsistent structural priors that limit realistic cellular organization. 

We propose \textbf{Dual-LoRA Controllable Diffusion}, a unified centroid-guided diffusion framework that jointly supports \emph{Local Structure Completion} and \emph{Global Structure Synthesis} within a single model. 
Multi-class nuclei centroids serve as lightweight and annotation-efficient spatial priors, providing biologically meaningful guidance under both partial and complete image absence. 
Two task-specific LoRA adapters specialize the shared backbone for local and global objectives without retraining separate diffusion models. Extensive experiments demonstrate consistent improvements over state-of-the-art GAN and diffusion baselines across restoration and synthesis tasks. 
% \pp{Some quantitative numbers?} 
For local completion, LPIPS computed within the masked region improves from 0.1797 (HARP) to 0.1524, and for global synthesis, FID improves from 225.15 (CoSys) to 76.04, indicating improved structural fidelity and realism.
Our approach achieves more faithful structural recovery in masked regions and substantially improved realism and morphology consistency in full synthesis, supporting scalable pan-cancer histopathology modeling.

% Authors must provide keywords and are not allowed to remove this Keyword section.

\end{abstract}
\section{Introduction}
\label{sec:intro}

Whole-slide imaging (WSI) enables high-resolution visualization of tissue specimens and 
has become a cornerstone of modern cancer diagnosis and computational pathology~\cite{janowczyk2016deep,cheng2020computational,pantanowitz2011review}. 
However, real-world WSIs exhibit substantial variability in staining protocols, tissue architecture, and cell density across organs and cancer types. 
They are further affected by artifacts such as stain inconsistency, blur, folds, foreign objects, and missing regions~\cite{seoane2004artefacts,taqi2018review}. 
These degradations often occur in diagnostically critical areas and complicate both expert interpretation and downstream computational analysis.

\begin{figure}[t]
    \centering
    \includegraphics[width=\linewidth]{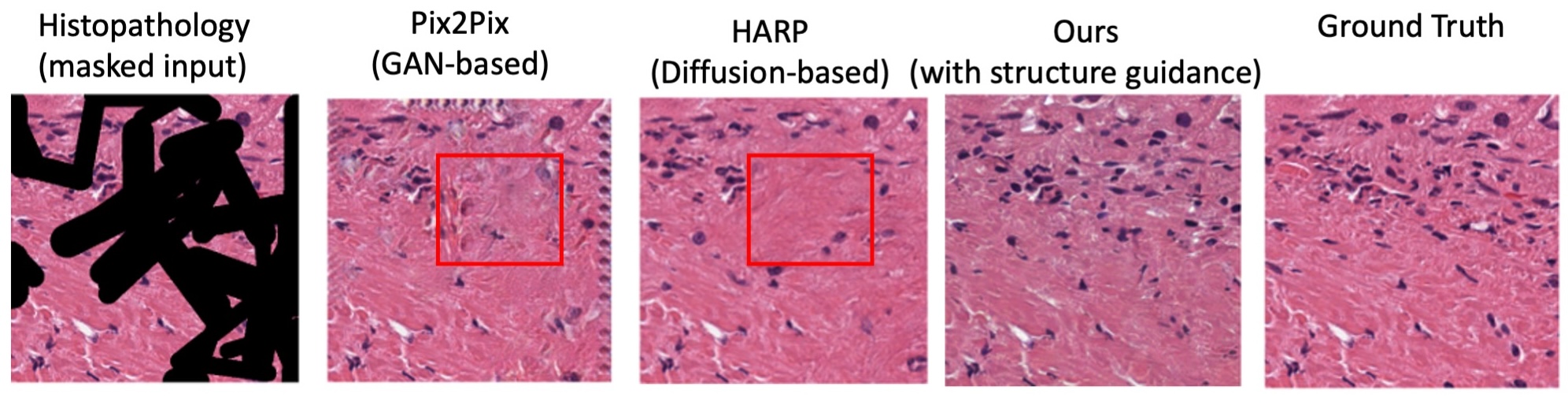}
    \caption{
    Challenges in histopathology local structure completion with complex masks.
    Pix2Pix~\cite{isola2017image} and HARP~\cite{harp} fail to reconstruct fine-grained structures under large irregular masks,
    while our method preserves coherent morphology and tissue continuity.
    Red boxes highlight reconstruction failures in the baselines.
    }
    \label{fig:intro_inpaint}
\end{figure}

Generative models have emerged as a promising solution for histopathology image restoration and data augmentation~\cite{jose2021generative,xu2025superdiff,li2022high}. 
Existing approaches address tasks such as artifact removal, stain normalization, cell layout generation, or patch-level synthesis~\cite{harp,rong2023enhanced,vitdae,xu2025topocellgen}. 
Despite encouraging results, most methods rely primarily on pixel-level inputs or global embeddings, lacking explicit spatial guidance of cellular organization~\cite{zoomldm,harp}. 
As a result, when reconstructing large missing regions or synthesizing tissue from limited cues, they may generate morphologically implausible structures (~\cref{fig:intro_inpaint}).
In practice, histopathology synthesis tasks typically arise in two closely related scenarios depending on the extent of missing tissue:
\textbf{(1) Local Structure Completion}, where partially missing tissue regions are reconstructed while preserving the surrounding histomorphology, and  
\textbf{(2) Global Structure Synthesis}, where an entire tissue patch is generated solely from structural priors when image content is fully absent.
Although typically treated as separate problems, both share a common objective: generating morphologically realistic H\&E tissue consistent with a biologically meaningful structural prior.

We argue that explicit spatial priors are essential for realistic synthesis in computational pathology. 
Dense annotations such as nuclei segmentation masks provide detailed structural information but are costly to obtain at scale~\cite{abdelsamea2022survey}. 
Self-supervised embeddings capture global context yet lack fine-grained cellular layout cues~\cite{zoomldm}. 
In contrast, \emph{cell-type centroids} offer a lightweight and annotation-efficient representation of nuclei location, density, and coarse cell identity. 
They encode biologically meaningful spatial organization while remaining scalable across pan-cancer datasets. 
Diffusion models naturally support flexible conditional generation and spatial control, making them well-suited for integrating such structured priors in histopathology synthesis~\cite{sd,controlnet}.

Building upon this observation, we present \textbf{Dual-LoRA Controllable Diffusion}, a unified centroid-guided diffusion framework that supports both local structure completion and global structure synthesis within a single model. 
A shared morphology-aware backbone provides consistent structural reasoning across cancer types, while two lightweight LoRA adapters~\cite{lora,xu2024ctrlora} specialize the model for the distinct optimization characteristics of the two tasks. 
This design enables task-specific adaptation without training separate diffusion models, promoting parameter efficiency and cross-task knowledge sharing.

% \pp{There were comments from reviewers regarding this. Have you addressed this somewhere with numbers now?} 

We train and evaluate our framework on a large-scale pan-cancer dataset spanning 30+ cancer types~\cite{tcga}, covering diverse organs, stain variations, and tissue architectures. 
Unlike prior work that typically focuses on a single cancer type or low-resolution patches (such as $128 \times 128$ or $256 \times 256$)~\cite{harp,min2024co}, our model operates at a resolution of $512 \times 512$ and jointly supports both local structure completion and global structure synthesis within a unified framework. 
Across both tasks, Dual-LoRA Controllable Diffusion consistently outperforms state-of-the-art GAN- and diffusion-based baselines under generic image quality metrics as well as pathology-aware evaluations. 
Downstream classification experiments further demonstrate improved preservation of discriminative histomorphology across cancer types.

Our contributions are summarized as follows:

\begin{itemize}
    \item \textbf{Centroid-guided unified diffusion backbone.} 
    We propose a structure-aware generative framework that jointly supports local structure completion and global structure synthesis across 30+ cancer types using multi-class nuclei centroids as scalable spatial priors.

    \item \textbf{Dual-LoRA specialization within a unified framework.} 
    Two lightweight LoRA adapters enable task-specific optimization for local structure completion and global structure synthesis while sharing a common morphology-aware backbone.
    
    \item \textbf{Comprehensive evaluation across restoration and synthesis tasks.} 
    Extensive experiments demonstrate consistent improvements over strong GAN and diffusion baselines on both inpainting and full synthesis tasks under generic and pathology-aware metrics.
\end{itemize}

\section{Methods}
\label{sec:methods}

\begin{figure*}[t]
  \centering
  \includegraphics[width=0.75\linewidth]{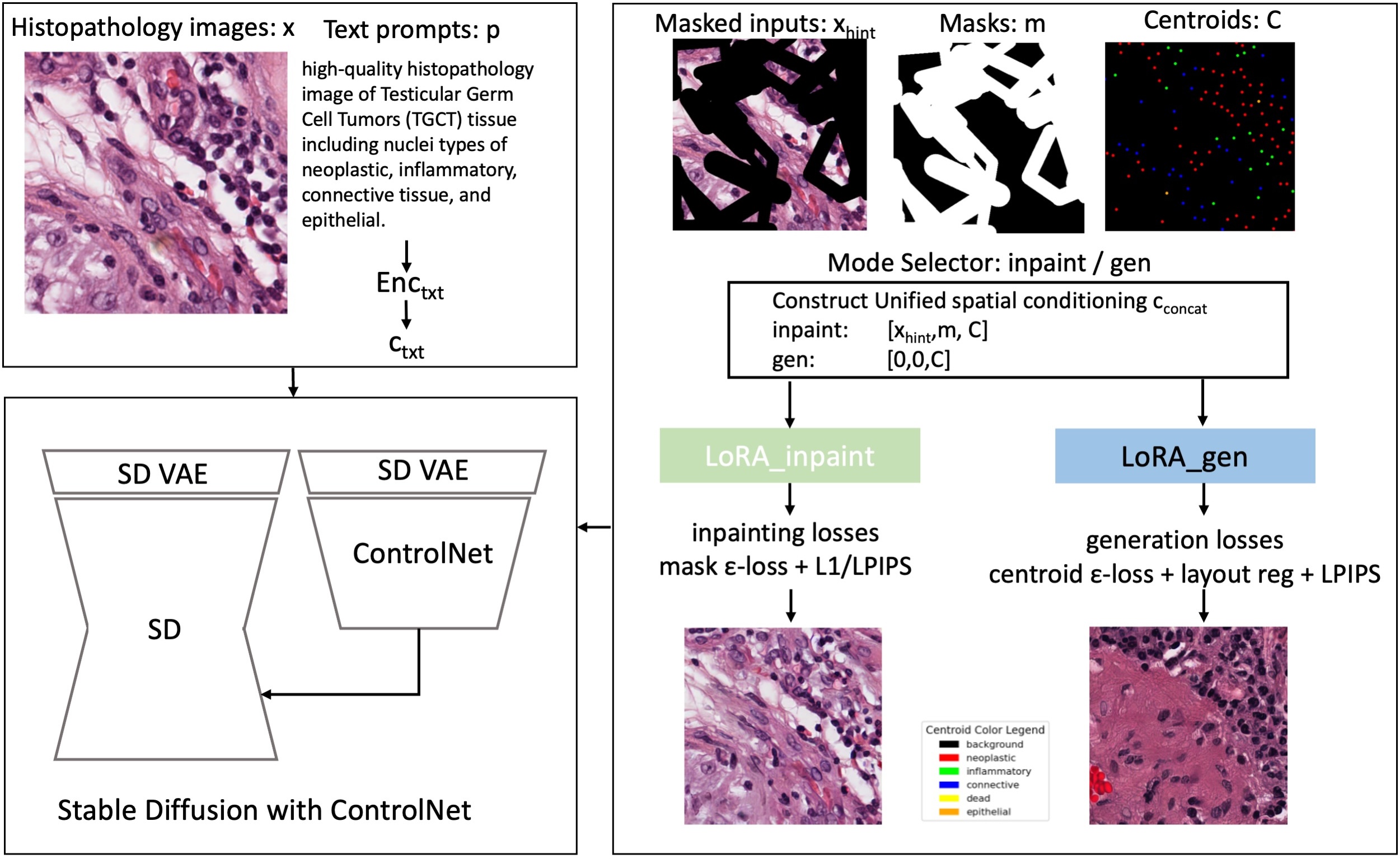}

% \caption{\textbf{Overview of Dual-LoRA Controllable Diffusion.}
% Given an input image $x$, mask $m$, and $K$ centroid maps $C$, we construct a unified spatial condition $c_{\mathrm{concat}}$. 
% For inpainting, $c_{\mathrm{concat}}^{(\mathrm{inpaint})}=[x_{\mathrm{hint}}, m, C]$ with $x_{\mathrm{hint}}=x\odot(1-m)$. 
% For generation, $c_{\mathrm{concat}}^{(\mathrm{gen})}=[\mathbf{0}, \mathbf{0}, C]$. 
% The Stable Diffusion backbone and ControlNet remain frozen, while two LoRA adapters specialize the model for the two synthesis modes. \pp{Can this caption be simplified?}}
\caption{Overview of Dual-LoRA Controllable Diffusion.
Given image $x$, mask $m$, and centroid maps $C$, we build a unified spatial condition.
For inpainting, $[x_{\mathrm{hint}}, m, C]$ with $x_{\mathrm{hint}}=x\odot(1-m)$; 
for generation, $[\mathbf{0}, \mathbf{0}, C]$. 
Stable Diffusion and ControlNet are frozen, while two LoRA adapters specialize the two modes.}
  \label{fig:overview}

\end{figure*}

Our proposed \textbf{Dual-LoRA Controllable Diffusion}, a centroid-guided latent diffusion framework, unifies two synthesis modes: 
(i) \emph{Local Structure Completion} and 
(ii) \emph{Global Structure Synthesis}. 
Both modes share a frozen Stable Diffusion backbone with ControlNet-based spatial conditioning~\cite{controlnet}, 
while two lightweight LoRA adapters enable task-specific specialization within a single architecture (~\cref{fig:overview}).

\subsection{Unified Problem Formulation}

Let $x \in \mathbb{R}^{H\times W\times 3}$ denote an H\&E image of spatial resolution $H \times W$, 
$C \in \mathbb{R}^{H\times W\times K}$ denote $K$ cell-type centroid maps, 
$m \in \{0,1\}^{H\times W}$ a binary mask indicating missing regions, and 
$p$ a text prompt encoding high-level tissue semantics. 
% \pp{Are all notations in the paper clearly defined? Please check carefully every single notation}

We unify both tasks through a single spatial control tensor:

\[
c_{\text{concat}}^{(s)} 
= [\, x_{\text{hint}}^{(s)},\, m^{(s)},\, C \,],
\quad s \in \{\text{inpaint},\text{gen}\}.
\]

For inpainting,
\[
x_{\text{hint}}^{(\text{inpaint})} = x \odot (1-m), 
\qquad 
m^{(\text{inpaint})} = m.
\]

For generation,
\[
x_{\text{hint}}^{(\text{gen})} = \mathbf{0}, 
\qquad 
m^{(\text{gen})} = \mathbf{0}.
\]

Thus, both modes share identical structural priors $C$, differing only in whether visual evidence is provided.

We follow latent diffusion~\cite{sd}, where $z_t$ is denoised as
\[
\epsilon_\theta =
\epsilon_\theta\big(
z_t,t;\,
c_{\text{concat}}^{(s)},
c_{\text{txt}};
\phi^{(s)}
\big).
\]
The text embedding $c_{\text{txt}}=\mathrm{Enc}_{\text{text}}(p)$ is obtained using the frozen CLIP encoder from Stable Diffusion~v1.5~\cite{clip,sd}, where $p$ encodes cancer-type and centroid semantics.

\subsection{Dual-LoRA Specialization}

Local completion emphasizes high-resolution texture recovery, 
whereas global synthesis requires large-scale spatial organization from centroids alone. 
Training a single parameterization for both leads to competing gradients.

Inspired by CtrLoRA~\cite{xu2024ctrlora}, we attach two lightweight LoRA adapters~\cite{lora} to the linear layers of the ControlNet~\cite{controlnet}, denoted as $\phi^{(\text{inpaint})}$ and $\phi^{(\text{gen})}$. 
For each backbone weight $W$, the effective weight becomes $W_{\text{eff}}^{(s)} = W + B^{(s)}A^{(s)}$, where only $A^{(s)}$ and $B^{(s)}$ are trainable. 
During training, only the active adapter receives gradient updates, preserving a shared morphology-aware representation while enabling task-specific modulation.

\subsection{Local Structure Completion}

The inpainting branch reconstructs masked tissue regions using both visible context and centroid priors. 
Centroids provide structural guidance inside the hole region, while visible tissue ensures boundary consistency.

\paragraph{Mask-aware diffusion loss.}

We emphasize masked regions during denoising by applying a spatially weighted $\epsilon$-prediction loss:
\[
\mathcal{L}_{\text{eps}}^{\text{inpaint}}
=
\| (\epsilon-\hat{\epsilon})
\odot w_{\text{mask}}
\odot w_{\text{snr}} \|_2^2,
\]
where $w_{\text{mask}}$ increases gradients inside missing regions and 
$w_{\text{snr}}$ follows the Min-SNR reweighting strategy~\cite{minsnr}.

\paragraph{Image-domain refinement.}

To improve perceptual fidelity within the hole, we decode the clean prediction $\hat{x}_0$ from latent space and introduce a softened mask $m_{\text{soft}}$ to restrict image-level supervision to the missing region:
\[
\mathcal{L}_{\text{img}}
=
\lambda_{L1}
\|(\hat{x}_0-x)\odot m_{\text{soft}}\|_1
+
\lambda_{\text{LPIPS}}
\,\mathrm{LPIPS}(\hat{x}_0,x).
\]
The perceptual term\cite{lpips} is activated after a warm-up phase to avoid destabilizing early diffusion training.

\paragraph{Final objective.}

\[
\mathcal{L}_{\text{inpaint}}
=
\mathcal{L}_{\text{eps}}^{\text{inpaint}}
+
\mathcal{L}_{\text{img}}.
\]

By combining masked observations with centroid-based structural priors, the model learns to reconstruct missing tissue while preserving cellular organization. 
The centroids reduce structural hallucination under large occlusions and promote anatomically consistent completions.
% \{-3.5mm}
\subsection{Global Structure Synthesis}

The generation branch synthesizes complete histopathology images from centroid priors and text semantics, without any observed pixels. 
Unlike inpainting, all structural information must be inferred from $C$, requiring explicit spatial guidance to ensure biologically plausible organization.

\paragraph{Centroid-aware diffusion loss.}

We adopt an $\epsilon$-prediction loss with spatial weighting to emphasize structurally meaningful regions:
\[
\mathcal{L}_{\text{eps}}^{\text{gen}}
=
\|(\epsilon-\hat{\epsilon})
\odot w_{\text{snr}}
\odot w_{\text{cent}}\|_2^2,
\]
where $w_{\text{snr}}$ follows Min-SNR reweighting~\cite{minsnr}, and $w_{\text{cent}}$ assigns higher weights near centroid locations, encouraging coherent cellular organization during denoising.

\paragraph{Layout regularization.}

To further stabilize global structure, we attach a lightweight layout head that predicts centroid-activation maps $\hat{C} \in [0,1]^{K\times H\times W}$ from ControlNet features. 
We impose an inter-class separation penalty~\cite{xu2025topocellgen}:

\[
\mathcal{L}_{\text{inter}}
=
\frac{2}{K(K-1)}
\sum_{i<j}
\frac{\langle \hat{C}_i,\hat{C}_j\rangle}{HW},
\]

which discourages spatial overlap across cell types and promotes structurally consistent layouts.

% Implementation details of the layout head and weighting strategy are provided in the supplementary material.

\paragraph{Final objective.}

The overall generation loss combines diffusion supervision and structural regularization:

\[
\mathcal{L}_{\text{gen}}
=
\mathcal{L}_{\text{eps}}^{\text{gen}}
+
w_{\text{inter}}\mathcal{L}_{\text{inter}}
+
\lambda_{\text{LPIPS}}
\,\mathrm{LPIPS}.
\]

A very weak perceptual term is applied only after warm-up to suppress minor visual artifacts without overriding structural guidance.
% \{-3.5mm}
\section{Experiments and Results}

\subsection{Dataset}

We construct a large-scale pan-cancer histopathology dataset from TCGA~\cite{tcga}, including 214,030 training patches from 7,245 whole-slide images (WSIs) spanning 31 cancer types, and 2,343 test patches from 18 WSIs 
covering 7 representative cancer types. All patches are $512\times512$ at $40\times$ magnification. Nuclei are segmented using HoVer-Net~\cite{hover}. 

% \pp{Should we say somewhere - may be in the discussion/conclusion that for practical utility, centroids can be generated using frameworks like TopoCellGen - that preserves spatial layout while providing variations of the configuration?} 
For each nucleus we extract its class among five categories and its centroid, which are rasterized into $K=5$ centroid maps used as structural priors. 
Text prompts encode cancer type and nuclei-class semantics.

\subsection{Implementation Details}

We build upon Stable Diffusion v1.5~\cite{sd} with a frozen ControlNet backbone~\cite{xu2024ctrlora}. 
Two rank-256 LoRA adapters are trained while the backbone remains fixed, yielding $\sim$113M trainable parameters (<10\% of full model). Training runs for 150k steps on three RTX 8000 GPUs with alternating modes. 
% Full hyperparameters are provided in the Supplementary.

% \begin{table}
% \begin{center}
% \scalebox{0.75}{
% \begin{tabular}{c|ccc}
% \hline
% Split & \#patches & \#cancer types & \#WSI \\
% \hline\hline
% Train & 214030 & 31 & 7245 \\
% Test & 2343 & 7 & 18 \\
% \hline
% \end{tabular}}
% \end{center}
% \caption{
% Dataset statistics for training and testing.
% }

% \label{tab:data_information}
% \end{table}

\begin{table*}
\begin{center}
\scalebox{0.5}{
\begin{tabular}{c|c|ccccccc}
\hline
Method & Condition & 
\multicolumn{1}{c}{FID(full)$\downarrow$} &
\multicolumn{1}{c}{IS(full)$\uparrow$} &
\multicolumn{1}{c}{LPIPS-FSD(full)$\downarrow$} &
\multicolumn{1}{c}{LPIPS(mask)$\downarrow$} &
\multicolumn{1}{c}{UNI-LPIPS(mask)$\downarrow$} &
\multicolumn{1}{c}{SSIM(mask)$\uparrow$} &
\multicolumn{1}{c}{PSNR(mask)$\uparrow$} \\
\hline \hline
Pix2Pix (GAN-based )~\cite{isola2017image}& Masked input  & 66.21 & 2.23 & 0.7847 & 0.1538 & 0.4827&\hspace{0.25em}\textbf{0.5316} \textsuperscript{*}&\hspace{0.25em}\textbf{17.53}\textsuperscript{*}\\
HARP (Diffusion-based)~\cite{harp} & Masked input& 60.27 & 2.19 & \hspace{0.25em}0.4967\textsuperscript{*} & 0.1797 &0.4664 & \textbf{0.5247} & \textbf{16.29} \\
\hline
Ours (Inpaint) & Prompt+Masked input  & \textbf{49.10} &\hspace{0.25em} \textbf{3.51}\textsuperscript{*}  & 0.6296 & \textbf{0.1524} &\textbf{0.4414}&0.4923  &15.38  \\ 
Ours (Inpaint) & Prompt+ Masked input + Centroids  &\hspace{0.25em}\textbf{37.39}\textsuperscript{*}& \textbf{3.08} &\hspace{0.25em}\textbf{0.4800}\textsuperscript{*}  &\hspace{0.25em}\textbf{0.1432}\textsuperscript{*}  & \hspace{0.25em}\textbf{0.4321}\textsuperscript{*}  &0.5014&15.90
  \\ %cfg scale 2.0

\hline
\end{tabular}}
\end{center}
\caption{
Quantitative comparison for local structure completion.}
\label{tab:inpaint_results}
\end{table*}

\begin{figure*}[t]
    \centering
    \includegraphics[width=0.8\textwidth]{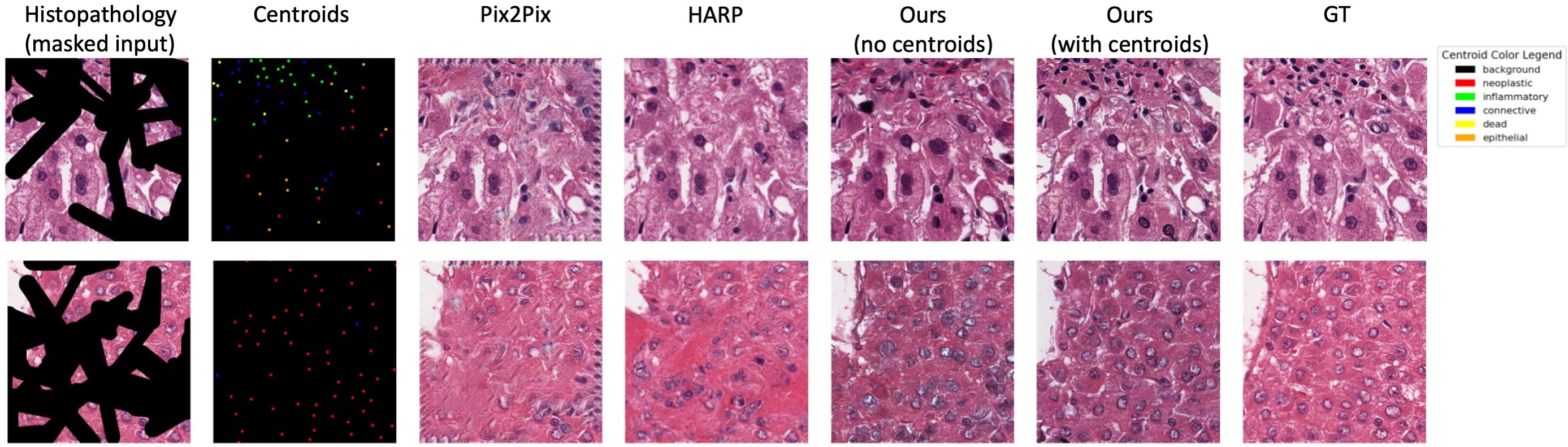}
\caption{
Qualitative comparison on inpainting under centroid-guided control.}

\label{fig:inpaint_results}
\end{figure*}

\begin{table*}
\begin{center}
\scalebox{0.8}{
\begin{tabular}{c|c|ccccc}
\hline
Method & Condition &
\multicolumn{1}{c}{FID$\downarrow$} &
\multicolumn{1}{c}{IS$\uparrow$} &
\multicolumn{1}{c}{LPIPS-FSD$\downarrow$} &
\multicolumn{1}{c}{UNI-FSD$\downarrow$} &
\multicolumn{1}{c}{UNI-LPIPS$\downarrow$} \\
\hline\hline
Pix2Pix (GAN-based)~\cite{isola2017image} & Centroids  & 98.45 & 1.76 & 2.20 & 331.15
& 0.9159 \\ 
CoSys (Diffusion-based)~\cite{min2024co} & Prompt + Centroids &  225.15 & 2.27 & 5.04 & 331.58 & 0.9448 \\ 
\hline
Ours (Gen) & Prompt + Centroids  & \textbf{76.04} &\textbf{3.26}  &\textbf{2.04} &\textbf{291.85} &\textbf{0.9055} \\ 
\hline
\end{tabular}}
\end{center}
\caption{
Quantitative comparison of centroid-guided global structure synthesis.
}
\label{tab:gen_results}
\end{table*}

\begin{figure}[t]
    \centering    \includegraphics[width=0.88\textwidth]{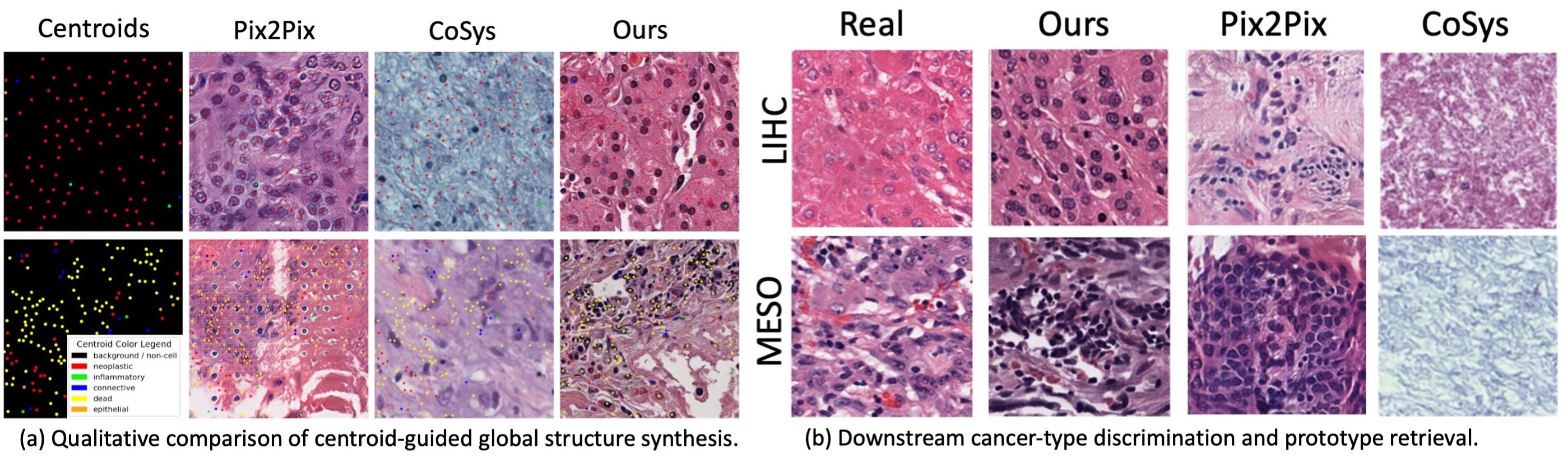}
\caption{
Qualitative evaluation of centroid-guided global structure synthesis.
(a) Under centroid guidance, Dual-LoRA produces coherent H\&E morphology compared to Pix2Pix and CoSys.
(b) Downstream cancer-type discrimination (LIHC, MESO) shows preservation of class-specific morphology.
}

\label{fig:centroid_res}
\end{figure}

\begin{table*}
\begin{center}
\scalebox{0.6}{
\begin{tabular}{c|c|cccc}
\hline
Method & Condition &
\multicolumn{1}{c}{Accuracy$\uparrow$} &
\multicolumn{1}{c}{Balanced Acc$\uparrow$} &
\multicolumn{1}{c}{Kappa$\uparrow$} &
\multicolumn{1}{c}{Weighted F1$\uparrow$} \\
\hline\hline
Pix2Pix (GAN-based)~\cite{isola2017image} & Centroids  & 0.5168 & 0.4717 & -0.0542 & 0.5249\\  

CoSys (Diffusion-based)~\cite{min2024co} & Prompt + Centroids & 0.8909 & 0.8822 & 0.7552 & 0.8915\\ 

\hline
Ours (Gen) & Prompt + Centroids  & \textbf{0.9506} &\textbf{0.9612}  &\textbf{0.8916} &\textbf{0.9513}  \\ 

\hline
\end{tabular}}
\end{center}
\caption{
Downstream cancer-type classification on synthetic LIHC and MESO patches.
 Our model outperforms Pix2Pix~\cite{isola2017image} and CoSys~\cite{min2024co} across all metrics.
}

\label{tab:gen_results_downstream}
\end{table*}

% \begin{figure}[t]
%     \centering
%     \includegraphics[width=0.45\textwidth]{figures/gen_results_downstream2.jpg}

% \caption{
% \textbf{Downstream cancer-type discrimination and prototype retrieval.}
% % Representative synthetic LIHC and MESO samples. Our method produces images
% % that align closely with the correct prototypes and preserve clear class-specific
% % morphology.
% Our synthetic LIHC and MESO samples align closely with the correct prototypes and preserve clear class-specific morphology.
% }
%   \label{fig:centroid_downstream}
% \end{figure}

\subsection{Local Structure Completion}
\paragraph{Quantitative results.}

Table~\ref{tab:inpaint_results} reports both full-image and mask-region metrics. 
In realistic settings where centroid annotations are unavailable at inference time, our model—trained with centroid-informed priors—already outperforms representative GAN-based methods such as Pix2Pix~\cite{isola2017image} and the state-of-the-art diffusion-based approach HARP~\cite{harp} in FID and LPIPS-based metrics With centroid guidance,
% \pp{ARe HARP and Pix2Pix SOTA? I thought this was a question by the reviewers?}. 
 performance further improves, achieving the best FID (37.39), LPIPS (mask) (0.1432), and UNI-LPIPS (mask) (0.4321).

Perceptual metrics consistently favor our method, while SSIM and PSNR are reported for completeness but correlate weakly with diagnostic realism~\cite{xu2025superdiff}.

\paragraph{Qualitative results.}

As shown in Fig.~\ref{fig:inpaint_results}, Pix2Pix~\cite{isola2017image} exhibits banding artifacts and HARP~\cite{harp} produces oversmoothed textures. 
Our model restores realistic stain variation and sharp nuclear boundaries, with centroid guidance improving structural coherence.

\subsection{Global Structure Synthesis}

\paragraph{Quantitative results.}

Table~\ref{tab:gen_results} summarizes centroid-guided full-image synthesis. 
Our method achieves the best performance across all generic and pathology-aware metrics. 
Compared to the recent diffusion-based baseline  CoSys~\cite{min2024co}, FID improves from 225.15 to 76.04 and LPIPS-FSD from 5.04 to 2.04, indicating substantial gains in realism and structural consistency. UNI-FSD and UNI-LPIPS computed using the UNI-2h encoder~\cite{uni2} further confirm improved histomorphology preservation.

% \paragraph{Topology-aware evaluation.}

% We evaluate spatial organization using TopoFD~\cite{xu2025topocellgen}. 
% Our method achieves the lowest TopoFD (0.000169), demonstrating superior preservation of higher-order nuclei topology (details in Supplementary).

\paragraph{Qualitative results.}

Fig.~\ref{fig:centroid_res} (a) shows that Pix2Pix~\cite{isola2017image} introduces artifacts and ignores layout, while CoSys~\cite{min2024co} often collapses into blurred regions. 
Our model produces coherent cellular organization aligned with centroid guidance.

\myparagraph{Downstream Cancer-Type Classification.} To assess whether synthetic images preserve discriminative morphology, we classify LIHC and MESO patches using UNI-2h embeddings~\cite{uni2} and ProtoNet~\cite{protonet} in a proof-of-concept study. 
As shown in Table~\ref{tab:gen_results_downstream}, Dual-LoRA achieves the highest balanced accuracy (0.96) and weighted F1 (0.95), improving over CoSys by approximately 9\% in balanced accuracy and 6.7\% in weighted F1. 
These results demonstrate that our framework best preserves cancer-type–specific histomorphology.
% \pp{Can we say by how much \% ?}. 

\section{Conclusion}
We presented \textbf{Dual-LoRA Controllable Diffusion}, a unified framework for structure-guided histopathology synthesis across pan-cancer data. 
By integrating centroid-based priors with a shared diffusion backbone and two lightweight LoRA adapters, the model supports both local completion and global synthesis within a single architecture. 
Experiments demonstrate improved realism and morphological fidelity over strong GAN and diffusion baselines. 
Centroid priors can be obtained from automated segmentation or layout-generation models such as TopoCellGen~\cite{xu2025topocellgen}, enabling scalable and controllable histopathology simulation. This flexibility makes the proposed framework scalable for data augmentation, simulation, and educational applications in computational pathology.
\bibliographystyle{splncs04}
\bibliography{mybibliography}
%
% \begin{thebibliography}{8}

% \bibitem{ref_article1}
% Author, F.: Article title. Journal \textbf{2}(5), 99--110 (2016)

% \bibitem{ref_lncs1}
% Author, F., Author, S.: Title of a proceedings paper. In: Editor,
% F., Editor, S. (eds.) CONFERENCE 2016, LNCS, vol. 9999, pp. 1--13.
% Springer, Heidelberg (2016). \doi{10.10007/1234567890}

% \bibitem{ref_book1}
% Author, F., Author, S., Author, T.: Book title. 2nd edn. Publisher,
% Location (1999)

% \bibitem{ref_proc1}
% Author, A.-B.: Contribution title. In: 9th International Proceedings
% on Proceedings, pp. 1--2. Publisher, Location (2010)

% \bibitem{ref_url1}
% LNCS Homepage, \url{http://www.springer.com/lncs}, last accessed 2023/10/25
% \end{thebibliography}
\end{document}